%% file: acl_latex.tex
\pdfoutput=1

\documentclass[11pt]{article}

\usepackage{acl}

\usepackage{xcolor}

\usepackage{booktabs}
\usepackage{graphicx}
\usepackage{times}
\usepackage{latexsym}
\usepackage{multirow}
\usepackage{numprint}
\usepackage[nameinlink]{cleveref}
\usepackage{pifont}
\usepackage{CJKutf8}

\usepackage[T1]{fontenc}

\usepackage[utf8]{inputenc}

\usepackage{microtype}

\title{NAVER LABS Europe's Multilingual Speech Translation Systems\\ for the IWSLT 2023 Low-Resource Track}

\author{Edward Gow-Smith$^1$\thanks{~~Work done during an
    internship at NAVER LABS Europe.} 
\quad Alexandre Bérard$^2$\thanks{~~Equal contribution} 
\quad Marcely Zanon Boito$^{2\dagger}$ 
\quad Ioan Calapodescu$^2$ 
\vspace{.2cm} \\ 
$^1$ University of Sheffield \qquad\qquad\qquad\hspace{.4cm} 
$^2$ NAVER LABS Europe \vspace{.1cm} \\
\tt{egow-smith1@sheffield.ac.uk}\quad
\tt{first.last@naverlabs.com}
}

\begin{document}
\maketitle
\begin{abstract}
{
This paper presents NAVER LABS Europe's systems for Tamasheq-French and Quechua-Spanish speech translation in the IWSLT 2023 Low-Resource track. 
Our work attempts to maximize translation quality in low-resource settings using multilingual parameter-efficient solutions that leverage strong pre-trained models.
Our primary submission for Tamasheq outperforms the previous state of the art by 7.5~BLEU points on the IWSLT~2022 test set, and achieves 23.6 BLEU on this year's test set, outperforming the second best participant by 7.7~points. For Quechua, we also rank first and achieve 17.7 BLEU, despite having only two hours of translation data.
Finally, we show that our proposed multilingual architecture is also competitive for high-resource languages, outperforming the best unconstrained submission to the IWSLT 2021 Multilingual track, despite using much less training data and compute.
}

\end{abstract}

\input{sections/1_intro}
\input{sections/2_architecture}
\input{sections/3_resources}

\input{sections/4_evaluation}

\input{sections/5_zero_shot}

\section{Conclusion}\label{sec:conclusion}

In this paper we presented our parameter-efficient multilingual systems as submissions to the IWSLT~2023 Low-Resource Task in the Tamasheq-French and Quechua-Spanish language pairs. The architecture we propose has several advantages: it is computationally and data efficient, it allows the same model to do both speech-to-text and text-to-text translation (or transcription), it maximizes knowledge transfer to improve low-resource performance, and it has good zero-shot translation capabilities.
Our submissions reach a new state of the art performance, winning both speech translation challenges, especially for Tamasheq-French, where we outperform the previous state of the art by more than~7~BLEU points.

Future work will include a comprehensive evaluation of the ASR capabilities of our architecture, and the investigation of adapters inside the speech representation model. Moreover, when the speech representation model is frozen, a more in-depth analysis of the optimal layer is needed.

\section*{Acknowledgements}
This work was partially funded by the European Horizon 2022 project UTTER (Unified Transcription and Translation for Extended Reality), under grant agreement No 101070631.

\bibliography{custom}

\input{sections/0_appendix}

\end{document}

%% file: sections/1_intro.tex
\section{Introduction}

The vast majority of speech pipelines are developed for \textit{high-resource} languages, a small percentage of languages that have ample amounts of annotated data available~\cite{joshi2020state}. However, the assessment of systems' performance based only on high-resource settings can be problematic, since it fails to reflect the real-world performance these approaches will have in diverse and smaller datasets. Moreover, as around half of the world's languages are considered to be not only \textit{low-resource}, but also from oral tradition~(i.e., without a written form),  there is an urgent need for speech technology that can operate robustly in such \textit{low-resource settings}~\cite{bird2011bootstrapping}.
In this context, the \textit{IWSLT conference}\footnote{\url{https://iwslt.org/}} proposes low-resource speech translation~(ST) challenges that allow the speech community to realistically benchmark ST approaches using diverse and representative datasets. 
This paper describes NAVER LABS Europe's (NLE) submission to two of the language pairs from the IWSLT~2023~\cite{iwslt:2023} Low-Resource Track:~Tamasheq-French (\textit{Taq-Fr}) and Quechua-Spanish (\textit{Que-Es}). 

\input{figures/architecture}

Most successful approaches for tackling scenarios where ST data is scarce perform transfer learning across languages and modalities, leveraging multilingual pre-trained models for both speech and text~\cite{anastasopoulos-etal-2022-findings}. 
However, due to the large number of parameters of current Transformer-based \cite{vaswani2017attention} approaches, training such systems is computationally expensive and not accessible to everyone.
\textbf{NLE's submission focuses on a multilingual parameter-efficient training solution that allows us to leverage strong pre-trained speech and text models to maximize performance in low-resource languages.}


We present new SOTA results for the \textit{Taq-Fr}~pair~(17\,hours of training data) that represent a 57\% BLEU increase compared to the results achieved by~\citeauthor{9834099}~(IWSLT~2022 post-evaluation).\footnote{\url{https://www.clsp.jhu.edu/jsalt-2022-closing-presentations/}} 
This same system achieves 23.6 BLEU on the IWSLT~2023 test set, an improvement of 7.71 BLEU compared to the second best result submitted this year.
We also present SOTA results in the unconstrained setting for the \textit{Que-Es} pair~(2\,hours of training data), while maintaining most of the performance in the \textit{Taq-Fr} pair.
In addition, to showcase the usefulness of our parameter-efficient multilingual solution we evaluate it on the high-resource setting of the IWSLT~2021 Multilingual Task~\cite{anastasopoulos-etal-2021-findings}. We find that our approach outperforms the best IWSLT~2021 submission (FAIR, \citealp{tang-etal-2021-fst}), despite training considerably fewer parameters~(-64\%), and using substantially less training data and compute. 

This paper is organized as follows. We first describe the architecture and training settings of our multilingual ST systems in Section~\ref{sec:system}. We next list the resources we use in Section~\ref{sec:resources}. Section~\ref{sec:results} presents our results in both low and high-resource settings. Lastly, we highlight the zero-shot potential of our approach in Section~\ref{sec:zeroshot} and present our concluding remarks in Section~\ref{sec:conclusion}.

%% file: figures/architecture.tex
\begin{figure*}
\centering
\includegraphics[scale=0.25]{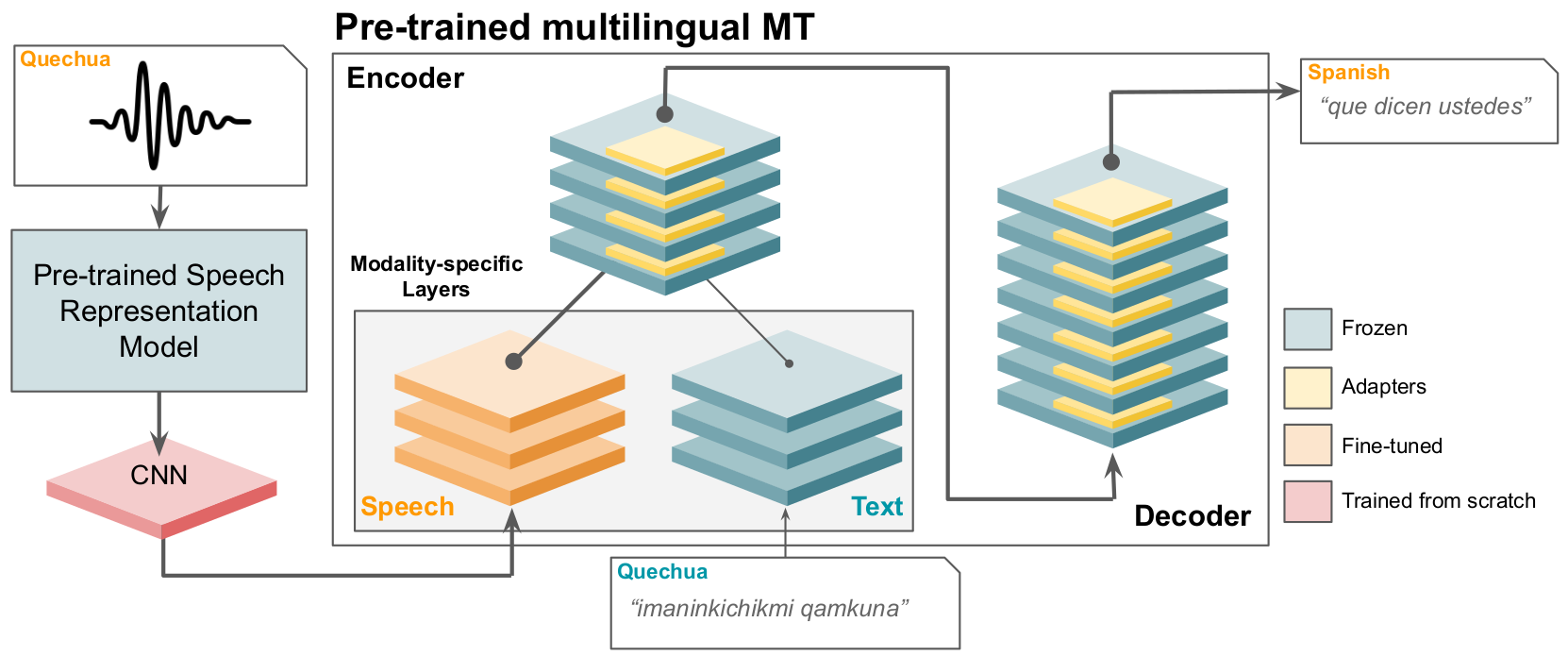}
    \caption{An illustration of our multilingual ST architecture as described in Section~\ref{sec:system}. The bold arrow path corresponds to the speech-to-text training path. At decoding time, we can choose between producing speech-to-text or text-to-text translations. Figure best seen in color.}
\label{fig:arch}
\end{figure*}

%% file: sections/2_architecture.tex
\section{System Description}\label{sec:system}

In this work we focus on a parameter-efficient training solution that allows us to input the features from a pre-trained speech representation model into a pre-trained multilingual MT model, producing translations from both speech and text in multilingual settings. This setting also allows us to leverage automatic speech recognition~(ASR; i.e. speech-to-transcript) data. The general architecture is presented in Figure~\ref{fig:arch}.
The architecture is considered \textit{parameter-efficient} because a small portion of its parameters are trained (bottom encoder layers and small adapters layers).

\paragraph{Architecture.}
We initialize our models with a pre-trained multilingual MT model, which we adapt to the ST task by inputting features extracted with a frozen pre-trained speech representation model. The MT model is also frozen, except for the bottom 2 or 3 encoder layers and small adapter modules~(those introduced by \citet{bapna-firat-2019-simple}, with bottleneck dimension 64) added after each encoder and decoder layer. As we show in our results, the fine-tuned encoder layers are able to map the speech features into the representation space of the pre-trained MT model and the adapters can help with domain adaptation~(and possibly help alleviate the length mismatch). At inference, this model can be used for MT with very little memory overhead: the convolutional layers and adapters are disabled, and the bottom encoder layers are swapped with those of the initial pre-trained model.

\paragraph{Training settings.}
We train on 4 V100 GPUs~(80GB) for up to \numprint{200000} updates, with a maximum batch size of \numprint{4000} source features (or 80 seconds of audio) and accumulated gradients over two batches.\footnote{This corresponds to a total of \numprint{32000} features per update, or 640 seconds of audio. In practice, with padding, each update corresponds to approximately 80 utterances or 530~seconds of audio.} We sample language pairs with a temperature of 3.\footnote{$p_{k} = u_k^{1/3}/\sum{u_{i}^{1/3}}$ where $u_k$ is the utterance count for language pair $k$.}
We validate every \numprint{5000} updates and perform early stopping on valid BLEU for the language pair(s) of interest, with a patience of 5, averaging model weights across the last 3 checkpoints.\footnote{While all the configurations presented in this paper use checkpoint averaging, we later re-trained our contrastive submission for \emph{Taq-Fr} and found virtually the same results without it.} We find best results using a single convolutional layer with stride 2, which downsamples the sequence of speech features by a factor of 2. The other hyperparameters are listed in Appendix Section~\ref{sec:app-hypeparameters}.

%% file: sections/3_resources.tex
\section{Resources}\label{sec:resources}

\subsection{Pre-trained Speech Representation Models}
We experiment with different versions of two speech representation models: HuBERT~\cite{9585401} and wav2vec~2.0~\cite{baevski2020wav2vec}. We do not fine-tune these models in any of our configurations, but instead use them as feature extractors~(see Figure~\ref{fig:arch}). 
Because of this, our models are sensitive to the layer we extract features from. \citet{pasad2021layer} argue that, for wav2vec~2.0 models that are not fine-tuned on ASR, speech features from \textit{middle} layers tend to have a higher abstraction from the speech signal, which is beneficial to downstream tasks. The results from \citet{zanon-boito-etal-2022-trac} seem to confirm this observation holds for low-resource ST.
To the best of our knowledge, there is no similar investigation for HuBERT models.\footnote{We hypothesize that layer selection is less important for HuBERT architectures due to the multi-iteration approach that increases signal abstraction at each iteration.}

\input{tables/models_speech}

Table~\ref{tab:models:speech} presents the speech representation models we experiment with. The \textit{Tamasheq} model is a monolingual wav2vec~2.0 Base model trained on 243\,h of Tamasheq speech. 
The \textit{Niger-Mali} is a wav2vec~2.0 Base model trained on the same Tamasheq speech data plus \numprint{111}\,h~of French, 109\,h~of Fulfulde, \numprint{100}\,h~of Hausa, and \numprint{95}\,h~of Zarma. This gives \numprint{658}\,h~in total. The data for both models is sourced from the Niger-Mali audio collection~\cite{zanon-boito-etal-2022-speech}.
The unreleased \textit{mHuBERT-Tamasheq} model uses this same audio collection for training, while also including Common Voice~\cite{ardila-etal-2020-common} data in four other languages~(English, French, Arabic and Kabyle), resulting in \numprint{5069}\,h~of speech. \textit{XLSR-53}~(56k~hours) and \textit{XLS-R}~(500k~hours) are massively multilingual wav2vec~2.0 Large models covering 53 and 128~languages, respectively. Neither of these two multilingual models have seen Tamasheq or Quechua speech during training.\footnote{Appendix Table~\ref{tab:models:speechurl} lists all models with links for downloading checkpoints, when available.}

\subsection{Pre-trained Multilingual MT Models}

To initialize our ST models, we first experimented with mBART for many-to-many translation (mBART50NN; \citealp{tang2020multilingual}), but found the NLLB-200 models \cite{costa2022no} to give better results. We experiment with the dense NLLB models of various sizes: the distilled 600M-parameter and 1.3B-parameter versions, and the 3.3B-parameter version. We end up using the larger versions in our submissions (1.3B and 3.3B).
Note that NLLB covers 202 languages, including Tamsheq and Quechua, which is not the case for mBART. At the same model size, despite covering more languages, NLLB is also a stronger machine translation model overall than mBART. Also, unlike mBART, it is not English-centric.

Contrary to \citet{tang-etal-2021-fst}, we keep the original mBART or NLLB vocabularies of size 250k and do not train any embeddings. Instead, like \citet{berard-etal-2021-efficient}, we find that it is possible to filter the vocabulary at test time to only cover the languages of interest, significantly reducing the memory footprint of the model with a minor reduction in performance.\footnote{With NLLB, 44k tokens are enough for a 100\% coverage of the training data (mTEDx, TED-LIUM, Quechua, Tamasheq), or 35k when restricting to our \textit{Taq-Fr} setting. This represents a reduction of more than 200M parameters.} We can also filter the vocabulary and embeddings before ST fine-tuning and achieve the same performance as with the full vocabulary without needing to train any embeddings. See Table~\ref{tab:vocab_filtering} in Appendix for a comparison of these approaches. In order to study the zero-shot translation capabilities of our models~(i.e., translating to languages and language pairs unseen at training), we do not apply vocabulary filtering to the configurations presented in the main paper.

\input{tables/data_LR}
\input{tables/data_HR_1}

\subsection{Datasets}
We tackle the low-resource setting by building multilingual systems that utilize both ASR and ST data in the languages of interest~(Tamasheq and Quechua), and in high-resource directions whose target language is of interest (French and Spanish). Note that we also include X$\to$English data, as we initially planned to participate in the Irish-English task.
Including more data in high-resource languages has several advantages. Firstly, it has a regularization effect that prevents us from immediately overfitting the low-resource training data. Secondly, this enables knowledge transfer from common target languages and from similarly-sounding source languages.\footnote{Manual inspection revealed that audio from both datasets presents some degree of target language borrowing (e.g., Spanish words present in the Quechua speech, French words present in the Tamasheq speech).} 
Thirdly, as we build multilingual ST systems by mapping the speech representation vectors into the same space as the multilingual MT model, our goal is to produce a model that is \textit{as multilingual as possible}, not specializing in one specific language. Our results show that training on multiple languages at once achieves this effect, while also producing good zero-shot ST results.

Table~\ref{tab:resources:lr} presents statistics for the datasets provided by the IWSLT~2023 organizers. The 
\textit{Que-Es} dataset\footnote{We are aware the dataset reference is \textit{Que-Spa}. We chose to use the ISO 639-1 two letters abbreviation for Spanish for consistency with the other datasets used in this work.} is an unreleased dataset prepared for this year's challenge. It corresponds to a translated subset of the Quechua ASR data~(``Siminchik'') from~\citet{cardenas2018siminchik}. The \textit{Taq-Fr} dataset was introduced by~\citet{zanon-boito-etal-2022-speech}. Table~\ref{tab:resources:highresource1} presents statistics for the datasets in high-resource languages. English ASR data comes from TED-LIUMv2~\cite{rousseau-etal-2014-enhancing}, and the other data comes from mTEDx~\cite{salesky2021mtedx}. Appendix Table~\ref{tab:data:check} lists the datasets used in each of our submissions.
In Section~\ref{sec:iwslt2021}, we also run experiments in the setting of the IWSLT~2021 Multilingual Task to measure how good our approach is on high-resource languages. The datasets used for this setting are presented in Appendix Table~\ref{tab:appendix:mtedx}.

%% file: tables/models_speech.tex
%
\begin{table}[]
\resizebox{\columnwidth}{!}{
\begin{tabular}{lccc}\bottomrule
\textbf{Model}                  & \multicolumn{1}{l}{\textbf{\# params}} & \textbf{\begin{tabular}[c]{@{}c@{}}Transformer \\ layers\end{tabular}} & \textbf{\begin{tabular}[c]{@{}c@{}}Feature \\ dimension\end{tabular}} \\\midrule
\textbf{Tamasheq}~\cite{zanon-boito-etal-2022-trac}               & 95M                                        & 12                         & 768 \\
\textbf{Niger-Mali}~\cite{zanon-boito-etal-2022-trac}        & 95M                                        & 12                         & 768                         \\
\textbf{mHuBERT-Tamasheq}            & 95M                                        & 12                         & 768                         \\
\midrule
\textbf{XLSR-53}~\cite{conneau21_interspeech}                & 317M                                       & 24                         & 1024                        \\
\textbf{XLS-R}~\cite{babu22_interspeech}               & 317M                                       & 24                         & 1024                        \\

\bottomrule         
\end{tabular}}
\caption{Speech representation models. The top portion presents \textit{Tamasheq-dedicated} models, while the bottom lists large \textit{general purpose} multilingual models.}
\label{tab:models:speech}
\end{table}

%% file: tables/data_LR.tex
\begin{table}[]
\scriptsize
\centering
\resizebox{\columnwidth}{!}{
\begin{tabular}{ccccc}\toprule
\textbf{Task} & \textbf{Source} & \textbf{Target} & \textbf{hours:minutes} & \textbf{\# utterances} \\\midrule
ASR           & Quechua         & Quechua         & 51:39                & 8,301                   \\\midrule
ST            & Quechua         & Spanish         & 2:42                 & 698                    \\
ST            & Tamasheq        & French          & 15:43                & 5,025                  \\
\bottomrule
\end{tabular}}
\caption{Speech Translation (ST) and Speech Recognition (ASR) data provided by the organizers (train+valid). The ASR data is outside of the constrained setting.}\label{tab:resources:lr}
\end{table}

%% file: tables/data_HR_1.tex
\begin{table}[]
\scriptsize
\resizebox{\columnwidth}{!}{
\begin{tabular}{ccccc}
\toprule
\textbf{Task} & \textbf{Source} & \textbf{Target} & \textbf{hours:minutes} & \textbf{\# utterances} \\\midrule
ASR           & English         & English         &         208:00               & 91,003                  \\
ASR           & French          & French          & 218:59              & 117,081                 \\
ASR           & Spanish         & Spanish         & 214:15              & 103,076                 \\
\midrule
ST            & French          & English         & 57:39               & 31,207                  \\
ST            & French          & Spanish         & 42:14               & 21,862                  \\
ST            & Spanish         & English         & 79:37               & 37,168       \\
ST            & Spanish         & French          & 9:34                   & 4,568                   \\
\bottomrule          
\end{tabular}}
\caption{ASR and ST data in English, French and Spanish sourced from TED talks (unconstrained setting).}\label{tab:resources:highresource1}
\end{table}

%% file: sections/4_evaluation.tex
\section{Experiments and Results}\label{sec:results}

All our submissions to the low-resource ST task are in the \textit{unconstrained} setting, due to the use of pre-trained models, and from training on data in other languages. The datasets used in each submission are listed in Appendix Table~\ref{tab:data:check}.
This section is organized as follows. We present our \textit{Taq-Fr} results~(\ref{sec:taqfraresults}) with a detailed ablation study justifying our architectural choices. We then present our \textit{Que-Es} results~(\ref{sec:quesparesults}). Lastly, we evaluate and analyze our approach in a high-resource setting (\ref{sec:iwslt2021}).

\subsection{Tamasheq-French Results}\label{sec:taqfraresults}

We submit two systems that have \textit{Taq-Fr} as the only low-resource language pair~(\textbf{primary} and \textbf{contrastive 1}). Additionally, we take our primary submission for \textit{Que-Es}, which has also been trained on \textit{Taq-Fr}, and submit this as \textbf{contrastive~2}. 
The top portion of Table~\ref{tab:results:low:submittedtest} gives the test BLEU scores, and the top portion of Appendix Table~\ref{tab:results:low:submittedvalid} presents the valid BLEU scores. Table~\ref{tab:results:low:stats} shows statistics (average and standard deviation) over multiple runs when applicable.

\input{tables/results_submitted_test}

\paragraph{System description.}
The \textbf{contrastive~1} model uses as a speech feature extractor the \textit{Niger-Mali} wav2vec~2.0 model~(8\textsuperscript{th} layer). It was initialized with NLLB~1.3B, whose bottom 3 encoder layers were finetuned. We took three runs of this setting with different random seeds and picked the best performing one on the validation set (in terms of \textit{Taq-Fr} BLEU) as our contrastive submission. We then ensembled the three runs as our \textbf{primary} submission.
Finally, \textbf{constrastive~2} is the ensemble model used as primary submission to the \textit{Que-Es} task, which covers both low-resource languages, and combines \emph{XSL-R Large} with \emph{NLLB 3.3B}.

\paragraph{Results.}
Our primary submission significantly outperforms the previous state of the art of 13.2~BLEU (+7.5~BLEU) on the IWSLT~2022 test set by~\citet{9834099}.\footnote{Here we are referencing the model pre-trained using the Niger-Mali dataset that was presented at JSALT 2022: \url{https://www.clsp.jhu.edu/jsalt-2022-closing-presentations/}} It also ranks first in this year's edition, with +7.7 BLEU over the second best primary submission. Our contrastive submissions rank second and third (beating the second best primary submission by +5.4 and +2.8 BLEU).

\subsubsection{Ablation Study}
\label{sec:ablation_study}

In Appendix Table~\ref{tab:ablation_study} we compare our \textbf{contrastive~1} model (the non-ensembled version of our primary submission) with other architectures trained on the same data to validate our choice of hyperparameters. 

\paragraph{Speech features.} The wav2vec 2.0 models trained with Tamasheq (\emph{Niger-Mali} and \emph{Tamasheq}) largely outperform the well-known massively multilingual models (\emph{XLSR-53} and \emph{XLS-R}) on \textit{Taq-Fr} (e.g. +2.5 BLEU \emph{Tamasheq} compared to \emph{XLS-R L}). These models are larger and trained on considerably more data, but do not include any Tamasheq speech.
Similar to previous works~\cite{pasad2021layer,zanon-boito-etal-2022-trac}, when extracting features from wav2vec 2.0  we find that the 8\textsuperscript{th} layer gives better results than the 11\textsuperscript{th} (penultimate) layer (+2.5 BLEU for \emph{Niger-Mali}).

For HuBERT, on the contrary, features from the 11\textsuperscript{th} layer give the best results (+0.2 BLEU compared to 8\textsuperscript{th} layer).
When using the \textit{right layer}, we find that wav2vec 2.0 outperforms HuBERT (+2.7 BLEU \emph{Niger-Mali} compared to \emph{mHuBERT-Taq}).

Finally, \emph{Niger-Mali} is as good on \textit{Taq-Fr} as the \emph{Tamasheq} wav2vec~2.0, but performs considerably better on \emph{Fr-En} (+4.1 BLEU), probably because it was trained with French audio. The best \emph{Fr-En} performance is achieved with \emph{XLS-R L}. We find worse performance on \emph{Fr-En} with \emph{XLS-R XL} (-2.0 BLEU), but this may be due to layer selection. 

\paragraph{Pre-trained MT model.} The larger the model used for initialization, the better the performance~(even more so for \textit{Fr-En}). However, we find that the gain from using NLLB 3.3B over NLLB~1.3B is too small to justify the increase in model size and decoding latency (3 times slower). At the same model size, NLLB 600M performs considerably better than mBART (+1.7 BLEU on \emph{Taq-Fr}, +3.6 BLEU on \emph{Fr-En}).

\paragraph{Trained parameters.} Fine-tuning too many encoder layers results in overfitting, which hurts \textit{Taq-Fr} and \textit{Fr-En} performance. On the other hand, fine-tuning just 1 or 2 layers instead of 3 does \emph{not} result in a large BLEU drop. Similarly, adapter modules are not always needed. Disabling decoder adapters does not degrade \textit{Taq-Fr} performance (+0.2 BLEU), but results in a slight drop in \textit{Fr-En} performance~(-0.9 BLEU), which could be attributed to a domain adaptation effect (to the mTEDx domain). Disabling encoder adapters has more impact on performance for \textit{\mbox{Taq-Fr}}~(\mbox{-0.8}~BLEU), with similar effect on performance for \textit{Fr-En} (-1.0 BLEU). Section~\ref{sec:iwslt2021} shows that these adapters are important for domain adaptation.

\paragraph{Convolutions.} The number of convolutional layers does not impact performance much (range of 1.1 BLEU on \textit{Taq-Fr} and 3.2 BLEU on \textit{Fr-En} for 0 to 3 layers), but it can have a large impact on decoding speed: each layer divides the input length by a factor of 2 resulting in a roughly 3.5$\times$ speed-up from 0 to 3 layers. Interestingly, even though it was trained on much shorter sequences, the MT model seems to adapt quite well to any input length, even without any convolutions -- we achieve a better \textit{Taq-Fr} result without any convolutions, but a worse \textit{Fr-En} result.\footnote{Without any convolution, the speech feature to target token ratio is \textbf{12:1}.} 
However, models with fewer convolutional layers seem to converge faster (as shown in Appendix Figure~\ref{fig:bleu_by_conv}).

\paragraph{Stacked layers.} While our approach described in Section~\ref{sec:system} fine-tunes some parameters of the pre-trained MT model, we can instead plug new Transformer layers at the bottom of the encoder, without changing any existing parameter.
These ``stacked layers'' result in slightly larger models but are conceptually simpler, as they try to map the speech features into the same representation space as the input text embeddings of the MT model.
Appendix Table~\ref{tab:stacked_layers} compares this architecture with the one used in our submission to the \emph{Taq-Fr} task.
We see that it performs similarly well (sometimes better) and that it does not add any noticeable decoding latency. We can even reach the same \emph{Taq-Fr} performance as our contrastive submission by just adding a single Transformer layer plus one convolution layer and small adapters (28M trained parameters in total).
Finally, disabling all adapters only results in a small BLEU drop, suggesting that it is indeed possible to map the speech features into the text input space, with only one Transformer layer. This is surprising, considering that the input to this layer is 6 times as long as the target sequence on average.

\subsection{Quechua-Spanish Results}\label{sec:quesparesults}
The test and validation scores of our submissions to the \textit{Que-Es} task are reported in the second half of Table~\ref{tab:results:low:submittedtest} and~\ref{tab:results:low:submittedvalid}, respectively. Because these models are also trained on \textit{Taq-Fr} data, we additionally report their performance on that task.

\paragraph{System description.} As we do not have a speech feature extractor specialized to Quechua speech, our \textbf{contrastive~1} submission uses a massively multilingual wav2vec 2.0 model: XLS-R Large~(18\textsuperscript{th} layer). Compared to our Tamasheq submission, it is also initialized with a larger MT model~(NLLB~3.3B), which we found to perform better in this setting. The training settings are the same as for the Tamasheq models, except that we only fine-tune the bottom 2 encoder layers (instead of 3) and validate every \numprint{2500} updates, since this larger model tends to converge faster. Another difference is that we train on both Tamasheq and Quechua data~(in addition to the mTEDx and TED-LIUM data).
Like in our Tamasheq submission, we train 3 models with different random seeds and ensemble them as our \textbf{primary} submission. Our \textbf{constrastive~2} submission uses a single model with the same training settings, but starts from a smaller pre-trained MT model~(NLLB~1.3B).

\paragraph{Results.}
Our primary submission in the \textit{Que-Es} task also ranked first, with 17.7~BLEU on the official test set. The full ranking results were not communicated in time to this camera-ready. They will be made available later through the conference findings paper~\cite{iwslt:2023}.

\paragraph{Data contamination.}
We found shortly after our submission that all the audio files used in the official test and validation sets are also present in the ASR training data shared by the organizers for the unconstrained setting.
This means that our \textit{Que-Es} ST models are evaluated in an unrealistic setting, where they are tasked to translate Quechua utterances of which they already know the transcription into Quechua.
For this reason, we filtered the ASR data to remove all audio files also present in the validation and test sets for \textit{Que-Es}, and we re-trained models on this filtered data.\footnote{In the updated version, we use NLLB 1.3B by default instead of NLLB 3.3B, like for \textit{Taq-Fr}. Appendix Table~\ref{tab:results:low:submittedvalid} presents \textit{uncontaminated} results.} 
While our official submission results presented in Table~\ref{tab:results:low:submittedtest} use the ``contaminated'' dataset for comparison with the other submissions, we think any future comparison to our work should be done with the updated results in Appendix Table~\ref{tab:results:low:submittedvalid}. Note that similar care should be taken with the results of other participants.

\subsection{Results and Analysis in a High-Resource Setting}
\label{sec:iwslt2021}

\input{tables/iwslt2021}

The results of our ablation studies (Section~\ref{sec:ablation_study}) seem to indicate that our models are reasonably good on \emph{Fr-En} translation, even though we do early stopping and tune our hyper-parameters based on \emph{Taq-Fr} performance.
Here, we further investigate the performance of our approach on high-resource ST by training models in the setting of the IWSLT 2021 Multilingual Task~\cite{anastasopoulos-etal-2021-findings}. This task evaluates the performance of multilingual ST models in 4 \textit{training directions}, for which in-domain training data is provided, and 3 \textit{zero-shot directions}, for which no training data is provided.

We use \emph{XLS-R Large} as the speech feature extractor, experiment with both \emph{NLLB 1.3B} and \emph{NLLB~3.3B} as the MT model, and perform early stopping based on the average validation BLEU across the 4 official training directions.
We train our models on all the mTEDx language pairs that are not zero-shot, along with TED-LIUM (English ASR) and the Tamasheq and Quechua data~(see Table~\ref{tab:data:check}).
Note that the use of pre-trained models and English ASR means our models fall into the unconstrained setting.

Table~\ref{tab:iwslt_2021_setting} presents our results on this task, compared with the best unconstrained submission~(FAIR; \citealp{tang-etal-2021-fst}).\footnote{SacreBLEU signature \cite{post-2018-call}: \texttt{\scriptsize nrefs:1|\\case:mixed|eff:no|tok:13a|smooth:exp|version:2.1.0}} We find that both our models outperform FAIR's ensemble submission in the training directions, even though they require substantially less compute and data to train, and they are not ensembled. In the zero-shot directions, our NLLB 1.3B version performs worse than FAIR's ensemble, which is not surprising since they used training data for the zero-shot language directions (from other datasets), whilst we do not.\footnote{NLLB has been pretrained on these language pairs for MT, but we do not train on ST data for them.} We find that using the larger NLLB 3.3B model for initialization considerably improves our zero-shot results.

\subsubsection{Incremental Learning}

\input{tables/incremental_learning}

A limitation of our approach for low-resource ST is that we need to know in advance (when training the multilingual ST model) the set of low-resource languages to cover. Here, we show that it is possible to add a new low-resource language into an existing model without re-training it, similar to what has been previously done by \citet{berard-2021-continual} for text-to-text MT.
We train a model following the IWSLT 2021 setting presented above, but without any Tamasheq or Quechua data. Then, we attempt to adapt it to \textit{Taq-Fr} using four different approaches: \textbf{1)} adding adapters of dimension 64 in the bottom layers and training all adapters (including in the decoder layers and top encoder layers); \textbf{2)} adding adapters of dimension 256 in the bottom layers and fine-tuning all adapters; \textbf{3)} adding adapters of dimension 256 in the bottom layers and training only those; \textbf{4)} adding adapters of dimension 256 in the bottom layers and training both those and the convolutional layer.

We keep the same training settings as before, except that: we train on \textit{Taq-Fr} data only; we train only the parameters mentioned above; we validate more often (every \numprint{1000} updates); and we disable checkpoint averaging. Table~\ref{tab:incremental} shows the performance of these four incremental training methods, compared to training on the entire language set from scratch. Even though incremental training does not perform quite as well, it appears to be a viable option that can achieve decent results. Lastly, we highlight that our experiments were limited to these four incremental learning settings (without hyper-parameter search), and that better results may be obtained with other parameter-efficient adaptation methods, or with more regularization.

\input{tables/zeroshot_table}

\subsubsection{Multimodality and Domain Transfer}

Since our systems are initialized with an MT model, of which just a few encoder layers are modified, it is straightforward to use our ST models for text-to-text translation: we just need to store both the MT and ST bottom layers and route tokens through the MT ones~(see Figure~\ref{fig:arch}).
However, one question that remains is whether the ST adapters can be used for text-to-text decoding.

As an investigation of this, Appendix Table~\ref{tab:domain_transfer} measures the MT performance~(NLLB 1.3B) on the IWSLT 2021 test sets (same domain as the mTEDx training data) with and without the ST adapters. Surprisingly, we see that not only can we use these adapters for both text and speech modalities, but they actually improve the MT scores (+2.7 BLEU on average), even though they were only trained with ST and ASR data.
This suggests that the fine-tuned bottom layers are able to fully map the speech representations into the text representation space and that the adapters further improve performance by allowing domain adaptation of the MT model (which is hard to do at the very bottom layers).
Note that the encoder adapters seem to be the most important ones, which is consistent with the findings of \citet{cooper-stickland-etal-2021-multilingual} that adapting the encoder is the most effective strategy for domain adaptation.
Lastly, we highlight that adapting the MT model directly with MT data (mTEDx's transcriptions and translations) gives even better results~(+4.6 BLEU on average), but this cross-modality domain transfer is an interesting by-product of our parameter-efficient approach.

%% file: tables/results_submitted_test.tex
\begin{table}
\small
\centering
\begin{tabular}{cccc|c}\toprule
&& \multicolumn{2}{c|}{\textbf{Taq-Fr}} & \textbf{Que-Es} \\
&                        & \begin{tabular}[c]{@{}c@{}}\textbf{IWSLT} \\ \textbf{2022}\end{tabular} & \begin{tabular}[c]{@{}c@{}}\textbf{IWSLT} \\ \textbf{2023}\end{tabular} & \begin{tabular}[c]{@{}c@{}}\textbf{IWSLT} \\ \textbf{2023}\end{tabular}               \\\midrule
\multirow{3}{*}{\textbf{\begin{tabular}[c]{@{}c@{}}Taq-\\ Fr\end{tabular}}} & \textbf{primary}       & \textbf{20.75}                                        & \textbf{23.59}                                        & $$ \ding{55} $$                                                                   \\
& \textbf{contrastive 1} & 19.06                                                 & 21.31                                                 & $$ \ding{55} $$                                                                  \\
& \textbf{contrastive 2} & 18.58                                                 & 18.73                                                 & \textbf{17.74}                                                               \\\midrule
\multirow{3}{*}{\textbf{\begin{tabular}[c]{@{}c@{}}Que-\\ Es\end{tabular}}} & \textbf{primary}       & 18.58            & 18.73              & \textbf{17.74}                                                      \\
& \textbf{contrastive 1} & 16.84          & $$ \ding{55} $$           & 15.67                                                               \\
& \textbf{contrastive 2} & 16.21                                               & $$ \ding{55} $$                                                    & 15.25                                                 \\\bottomrule             
\end{tabular}
\caption{Results on the official test sets for the IWSLT 2023 Low-Resource Task. We 
also show results on the IWSLT 2022 \textit{Taq-Fr} test set. Note that all Quechua models are trained on Tamasheq data, but the reverse is not true~(see Appendix Table~\ref{tab:data:check}). Lines 3 and 4 correspond to the same model.}
\label{tab:results:low:submittedtest}
\end{table}

%% file: tables/iwslt2021.tex
\begin{table*}[]
\centering
\small
\begin{tabular}{c|c|c|cccc|ccc}\toprule
\multirow{2}{*}{\textbf{Model}} & \textbf{Total} & \textbf{Trained} & \multicolumn{4}{c|}{\textbf{Training directions}} & \multicolumn{3}{c}{\textbf{Zero-shot directions}} \\
& \textbf{params} & \textbf{params} & \textbf{Es-En} & \textbf{Fr-En} & \textbf{Fr-Es} & \textbf{Pt-En} & \textbf{Pt-Es} & \textbf{It-En} & \textbf{It-Es} \\
\midrule
FAIR at IWSLT 2021 & \multicolumn{2}{c|}{700M} & 40.4 & 36.4 & 34.4 & 29.0 & 34.4 & 28.4 & 34.6 \\
\cite{tang-etal-2021-fst} & \multicolumn{2}{c|}{3$\times$700M (ensemble)} & 42.2 & 38.7 & 36.5 & 31.0 & \textbf{38.2} & \textbf{29.4} & \textbf{37.3} \\
\hline
XLS-R + NLLB 1.3B & 317M + 1.38B & 70M & 43.7 & 39.4 & 38.0 & 31.5 & 35.9 & 28.9 & 35.0 \\
XLS-R + NLLB 3.3B & 317M + 3.36B & 115M & \textbf{44.0} & \textbf{39.9} & \textbf{38.3} & \textbf{33.1} & 38.1 & 29.3 & 36.9 \\
\midrule
\multicolumn{3}{c|}{XLS-R + NLLB 1.3B, ASR + MT cascade} & 41.8 & 35.6 & 34.4 & 29.7 & 35.8 & 29.3 & 35.2 \\ 
\bottomrule
\end{tabular}
\caption{Results on the IWSLT 2021 Multilingual task. We report BLEU scores on the IWSLT 2021 test sets. Our NLLB 1.3B and 3.3B models took respectively 34 and 46\,h to train on 4 V100 GPUs, while FAIR's models each took 7 days to train on 8 V100 GPUs. Also note that FAIR's models were trained on much larger amounts of data, \textbf{including data for the ``zero-shot'' directions} (which, in their case is only zero-shot w.r.t the in-domain TED data).}
\label{tab:iwslt_2021_setting}
\end{table*}  

%% file: tables/incremental_learning.tex
\begin{table}[]
    \centering
    \small
    \begin{tabular}{l|c|c}\toprule
         \textbf{Model} & \textbf{New params} & \textbf{Taq-Fr} \\
         \midrule
         Joint training & 0 & \textbf{21.06} \\
         \midrule
         Adapters 64 (all) & 6.4M & 17.60 \\
         Adapters 256 (all) & 15.9M & 18.18 \\
         Adapters 256 (bottom) & 1.6M & 19.24 \\
         Conv + Adapters 256 (bottom) & 2.5M & 19.13 \\\bottomrule
    \end{tabular}
    \caption{BLEU scores on the \textit{Taq-Fr} validation set, when training jointly with IWSLT 2021 and Tamasheq data; versus incremental (2-stage) training. The ``New params'' columns give the number of Tamasheq-specific parameters added.}
    \label{tab:incremental}
\end{table}

%% file: tables/zeroshot_table.tex
\begin{table*}[]
\small
\centering
\begin{tabular}{cccccccccc}\toprule
\textbf{\begin{tabular}[c]{@{}c@{}}Adapter \\ Size\end{tabular}} & \textbf{\begin{tabular}[c]{@{}c@{}}Encoder \\ Adapters\end{tabular}} & \textbf{\begin{tabular}[c]{@{}c@{}}Decoder \\ Adapters\end{tabular}} & \textbf{\begin{tabular}[c]{@{}c@{}}Taq-Fr\\ BLEU\end{tabular}} & \textbf{\begin{tabular}[c]{@{}c@{}}Taq-En\\ BLEU\end{tabular}} & \textbf{\begin{tabular}[c]{@{}c@{}}Taq-Ko\\ BLEU\end{tabular}} & \textbf{\begin{tabular}[c]{@{}c@{}}Taq-Fr\\ chrF\end{tabular}} & \textbf{\begin{tabular}[c]{@{}c@{}}Taq-En\\ chrF\end{tabular}} & \textbf{\begin{tabular}[c]{@{}c@{}}Taq-Ko\\ chrF\end{tabular}} \\\midrule
64 & \ding{51} & \ding{51} & 19.1 & \textbf{17.1} & 12.6 & 44.2 & 40.8 & 18.2 \\
128 & \ding{51} & \ding{51} & 19.2 & 16.7 & 9.6 & \textbf{44.7} & 40.3 & 14.5 \\
64 & \ding{51} & \ding{55} & \textbf{19.3} & 16.8 & 14.6 & 44.4 & \textbf{42.4} & 21.5 \\
\ding{55} & \ding{55} & \ding{55} & 17.5 & 16.2 & 14.4 & 43.0 & 40.8 & 21.5 \\\midrule
\multicolumn{3}{c}{ST (contrastive 1) + MT (NLLB 1.3B) cascade} & \ding{55} & 15.0 & \textbf{15.7} & \ding{55} & 38.6 & \textbf{22.2} \\
\bottomrule
\end{tabular}
\caption{BLEU and chrF results for \textit{Taq-\{Fr, En, Ko\}} using \textbf{contrastive 1} and its variants (models trained without adapters or with larger adapters), on the IWSLT~2022 \textit{Taq-Fr} test set or silver-standard Korean and English references obtained with MT. The last row is a cascade of speech translation followed by text translation (Taq$\to$Fr$\to$X).}
\label{tab:zeroshot:results}
\end{table*}

%% file: sections/5_zero_shot.tex
\section{Zero-Shot Capabilities}\label{sec:zeroshot}

Throughout this paper we have argued that one advantage of the multilingual models we propose is their potential for zero-shot translation, a setting in which a system produces translation in an unseen language pair by leveraging its existing knowledge of both languages. In Section~\ref{sec:iwslt2021} we showed that our models are competitive with the best submission to IWSLT~2021 on the three zero-shot high-resource language pairs, despite the fact that these pairs were not truly zero-shot for that system.
In this section, we further illustrate the zero-shot capabilities of our models by translating Tamasheq speech in two settings: \textbf{1)} target language seen during both MT pre-training and ST adaptation~(English); \textbf{2)} target language only seen during MT pre-training~(Korean).

\paragraph{Evaluation settings.} To score BLEU and chrF\footnote{SacreBLEU signature: \texttt{\small nrefs:1|case:mixed|\\eff:no|tok:X|smooth:exp|version:2.3.1}, (En: \texttt{\small X=13a}, Ko: \texttt{\small X=ko-mecab-0.996/ko-0.9.2-KO}). chrF signature: \texttt{\small nrefs:1|case:mixed|\\eff:yes|nc:6|nw:0|space:no|version:2.3.1}} in the chosen target languages, we use a commercial translation service to translate the French side of the IWSLT~2022 test set to English and Korean. 
Note that this is only a \textit{silver-standard} made of synthetic data, and thus the evaluation will inevitably be biased.\footnote{For instance, we observe that these generated translations contain both the Korean transliteration in Hangul of named entities and the original version in the Latin script. This will likely penalize our produced translation during scoring.} Our goal is solely to assess whether our systems have \textit{some} zero-shot ST abilities.
We evaluate our \textit{Taq-Fr} \textbf{contrastive 1} system, and variants of this system with fewer or larger adapters.
We compare with a \textit{cascade} baseline, in which we first perform \textit{Taq-Fr} ST, followed by \textit{Fr-En} or \textit{Fr-Ko} MT using the text-to-text path from Figure~\ref{fig:arch}. In this setting, the adapters are disabled during MT.

\paragraph{Results.} In Table~\ref{tab:zeroshot:results}, we measure the zero-shot translation capabilities of our approach on this silver-standard test set. We evaluate four models: our \textbf{contrastive 1} submission presented in Section~\ref{sec:taqfraresults}, and variants of this model with increased adapter size, adapters only in the encoder, or no adapters. We compare against a cascade baseline that is not zero-shot, which consists in translating the Tamasheq speech into French text and then translating this text into English or Korean.

We observe that, in the case of English, which was seen during ST adaptation, adapters can be helpful (+2 BLEU over the cascade baseline).
On the other hand, for Korean, unseen during ST adaptation, systems with adapters in the decoder~(first two rows) perform worse, as they likely bring some degree of \textit{language confusion}. Results are even worse with larger adapters, with over 40\% of output sentences being in the wrong language. In this setting, the best results are achieved with only encoder adapters or no adapters at all~(-1~BLEU compared to the baseline).

\input{tables/zeroshot_examples}

Appendix Table~\ref{tab:zeroshot:langid} measures the percentage of output sentences in the correct language and the percentage of Hangul versus Latin character in each system's outputs. We find that models with adapters in the decoder~(first two rows) generate more Latin characters. Note that the ideal translation is not necessarily 100\% Hangul, as it might sometimes be best to keep the foreign named entities in the Latin alphabet. Table~\ref{tab:zeroshot:examples} illustrates this with a few examples of translations from our \textbf{contrastive 1} system.

%% file: tables/zeroshot_examples.tex
 \begin{table*}
\scriptsize
\resizebox{\textwidth}{!}{
\begin{tabular}{cc|l}
\toprule
\multicolumn{1}{c}{\textbf{Utterance id}} & \multicolumn{1}{c}{\textbf{Target}} & \multicolumn{1}{l}{\textbf{Content}}   \\\midrule
\multirow{4}{*}{\textbf{2016-11-23\_id\_7}} & \textbf{Ref}                        & Chers auditeurs, rappelez-vous que vous écoutez \textbf{Studio Kalangou} en ce moment.\\
& \textbf{Fr}                         & Chers auditeurs, n'oubliez pas que vous êtes avec le \textbf{Studio Kalangou}.\\
& \textbf{En}                         & Well, listeners, don't forget that you are with \textbf{Studio Kalangou} right now. \\
& \textbf{Ko}                         & \begin{CJK}{UTF8}{mj}청취자 여러분, 지금 \textbf{Studio Kalangou}와 함께 있는 것을 잊지 마세요.\end{CJK} \\\midrule
\textbf{2016-06-27\_id\_5}                  & \textbf{Ref}                        & Les examens du \textbf{BEPC} sont terminés et les corrections ont commencé hier après-midi dans la ville de Niamey.\\
\textbf{}                                 & \textbf{Fr}                         & Les examens du \textbf{BEPC} sont terminés et sur toute l'étendue du territoire, les travaux de leur suivi ont débuté hier après-midi à Niamey.\\
\textbf{}                                 & \textbf{En}                         & The \textbf{BEPC} exams are over and throughout the country, the monitoring activities started yesterday afternoon in Niamey. \\
& \textbf{Ko}                         & \begin{CJK}{UTF8}{mj}\textbf{BEPC} 시험은 끝났습니다. 전국에서 검사 작업은 어제 오후 Niamey에서 시작되었습니다.\end{CJK}    \\  \midrule    
\multirow{4}{*}{\textbf{2016-10-27\_id\_39}}         & \textbf{Ref}                        & \begin{tabular}[c]{@{}l@{}}D'autres informations que nous apportons aujourd'hui concernent un projet appelé \textbf{aniamey.com} qui informe que l'État du Nigéria a refoulé \\ des Nigériens, au nombre de 53, qui arrivent (), qui habitent dans la ville de Mina sur le territoire du Niger ou Neja.\end{tabular} \\
& \textbf{Fr}                         & \begin{tabular}[c]{@{}l@{}}D'autres informations que nous apportons aujourd'hui concernent les informations apportées par un programme dénommé \textbf{Niamey Point Com} qui a \\ apporté des informations selon lesquelles le Nigeria a accueilli 53 Nigériens qui habitent la ville de Mena qui se trouve sur le territoire du Niger ou le Niger.\end{tabular} \\
& \textbf{En}                         & \begin{tabular}[c]{@{}l@{}}Today, we're going to talk about the information about a program called \textbf{Niamey Point Com}, which reports that Nigeria has brought back 53 Nigerians \\ who live in the town of Mena in Niger.\end{tabular}  \\
& \textbf{Ko}                         & \begin{CJK}{UTF8}{mj}우리게임의 오늘 기사에서는 \textbf{Niamey Point Com}라는 프로그램으로 나이지리아가 미네에 거주하는 53명의 니그르인을 귀환시켰다는 소식이 있습니다.\end{CJK} \\\bottomrule
               
\end{tabular}}
\caption{Some decoding examples for \textit{Taq-Fr}, \textit{Taq-En} and \textit{Taq-Ko} language pairs, accompanied by the French reference~(Ref). Utterance id corresponds to the suffix of the audio files in the IWSLT~2022 test set.}
\label{tab:zeroshot:examples}
\end{table*}

%% file: sections/0_appendix.tex
\clearpage

\appendix

\section{Appendix}\label{sec:appendix}

\subsection{Hyperparameters}\label{sec:app-hypeparameters}

\input{tables/hyperparameters}

\subsection{Additional Results}

\input{tables/data_HR_mtedx_full}
\input{tables/results_submitted_valid}
\input{tables/zeroshot_langid}

\input{tables/vocab_filtering}

\input{tables/data_setup_check}
\input{tables/models_speech2}
\input{tables/stacked_layers}
\input{tables/ablation_study}

%% file: tables/hyperparameters.tex
\begin{table}[h!]
    \centering
    \scriptsize
    \resizebox{\columnwidth}{!}{
    \begin{tabular}{c|c}\toprule
         \textbf{Hyper-parameter} & \textbf{Value}  \\
         \midrule
         Batch size & \numprint{4000} \\
         Data-parallel GPUs & 4 \\
         Update freq & 2 \\
         Max learning rate & 0.0005 \\
         Initial LR & $10^{-7}$ \\
         Schedule & inverse square root \\
         Warmup steps & \numprint{10000} \\
         Adam betas & 0.9, 0.999\\
         Mixed precision & True \\
         Label smoothing & 0.2 \\
         Weight decay & 0.0 \\
         Dropout & 0.3$^\dagger$ \\
         Attention dropout & 0.1 \\
         Gradient clipping & none \\
         \midrule
         1D Convolutions & 1 \\
         Conv channels & 80$^\star$ \\
         Conv kernel size & 5 \\
         Conv stride & 2 \\
         Embed scaling factor & $\sqrt{\numprint{1024}}$\\
         Positional encoding & sinusoidal$^\alpha$ \\
         Encoder layers & 24 \\
         Decoder layers & 24 \\
         Embed dim & \numprint{1024}$^\ddagger$ \\
         FFN dim & \numprint{8192} \\
         Activation & ReLU \\
         Attention heads & 16 \\
         Pre-norm & True \\
         Adapter dim & 64 \\
         \midrule
         Vocab size & 250k \\
         Lang-pair temperature & 3 \\
         Heterogeneous batches & True\\
         Valid freq & \numprint{5000} \\
         Checkpoint averaging & 3 \\
         Patience & 5 \\
         Early stopping metric & BLEU \\
         Beam size & 5 \\\bottomrule
    \end{tabular}}
    \caption{Hyper-parameters used to train our models. \\$\star$: a linear layer followed by a ReLU activation is trained to project the input features (of dimension 768 or \numprint{1024}) to the input dimension of the CNN (80). \\$\dagger$: dropout is also applied to the source and target embeddings (after the convolutions and positional encoding) and FFN activations. \\$\ddagger$: \numprint{2048} when the pre-trained MT model is NLLB 3.3B. \\$\alpha$: learned positional embeddings in the decoder when the pre-trained model is mBART.}
    \label{tab:hyperparameters}
\end{table}

%% file: tables/data_HR_mtedx_full.tex
\begin{table}[h!]
\resizebox{\columnwidth}{!}{
\begin{tabular}{ccccc}
\toprule
\textbf{Task} & \textbf{Source} & \textbf{Target} & \textbf{hours:minutes} & \textbf{\# utterances} \\\midrule
ASR           & French          & French          & 218:59                 & 117,081                \\
ASR           & Italian         & Italian         & 118:39                 & 50,895                 \\
ASR           & Portuguese      & Portuguese      & 179:33                 & 91,257                 \\
ASR           & Spanish         & Spanish         & 214:15                 & 103,076                \\
\midrule
ST            & French          & English         & 57:39                  & 31,207                 \\
ST            & French          & Spanish         & 42:14                  & 21,862                 \\
ST            & French          & Portuguese      & 26:53                  & 14,322                 \\
ST            & Portuguese      & English         & 63:13                  & 31,868                 \\
ST            & Spanish         & French          & 9:34                   & 4,568                  \\
ST            & Spanish         & English         & 79:37                  & 37,168                 \\
ST            & Spanish         & Italian         & 11:50                  & 5,616                  \\
ST            & Spanish         & Portuguese      & 47:01                  & 22,012                \\
\bottomrule
\end{tabular}}
\caption{Statistics for all the mTEDx languages~(train+valid) seen by our systems for the IWSLT 2021 evaluation setup described in Section~\ref{sec:iwslt2021}.}
\label{tab:appendix:mtedx}
\end{table}

%% file: tables/results_submitted_valid.tex
\begin{table}[]
\resizebox{\columnwidth}{!}{
\begin{tabular}{llccc}\toprule
 &                        & \textbf{Taq-Fr valid} & \textbf{Que-Es valid} & \textbf{Que-Es test} \\\midrule
\multirow{3}{*}{\textbf{Taq-Fr}
} & \textbf{primary}       & \textbf{26.13}         & $$ \ding{55} $$  & $$ \ding{55} $$                  \\
 & \textbf{contrastive 1} & 24.53                  & $$ \ding{55} $$    &  $$ \ding{55} $$                 \\
 & \textbf{contrastive 2} & 22.88                  & \textbf{20.29}    &   \textbf{17.74}   \\\midrule
\multirow{3}{*}{\textbf{Que-Es}
} & \textbf{primary}       & 22.88                  & \textbf{20.29}      &  \textbf{17.74}  \\
 & \textbf{contrastive 1} & 20.81                  & 19.03            &   15.67     \\
 & \textbf{contrastive 2} & 21.31                  & 16.78         &     15.25 \\\midrule    
\multirow{3}{*}{\textbf{\begin{tabular}[c]{@{}l@{}}Que-Es\\ (updated)\end{tabular}}} & \textbf{primary} & 22.36 & 16.52 & \textbf{15.70}  \\
 & \textbf{contrastive 1} & 20.97 & 15.15 & 15.55 \\
 & \textbf{contrastive 2} & 20.31 & 16.30 & 13.17
 \\\bottomrule

\end{tabular}
}
\caption{Validation and test results on the IWSLT 2023 low-resource track. Lines 3 and 4 correspond to the same model. The ``\textit{Que-Es} (updated)'' results correspond to new models trained on filtered Quechua ASR data, where we removed audio files that are also in the ST valid and test sets. In this updated version, \textbf{primary} and \textbf{contrastive 1} use NLLB 1.3B and \textbf{contrastive 2} uses NLLB 3.3B.}
\label{tab:results:low:submittedvalid}
\end{table}

\begin{table}[]
\resizebox{\columnwidth}{!}{
\begin{tabular}{llcc}\toprule
&                        & \textbf{Taq-Fr test} & \textbf{Que-Es valid} \\\midrule
\multirow{2}{*}{\textbf{\begin{tabular}[c]{@{}l@{}}Taq-Fr\end{tabular}}} & \textbf{contrastive 1}       & 19.13 $\pm$ 0.06 & $$ \ding{55} $$                      \\
& \textbf{contrastive 2} & 16.89 $\pm$ 0.18 & 18.34 $\pm$ 0.59 \\\midrule
\textbf{\begin{tabular}[c]{@{}l@{}}Que-Es\end{tabular}} & \textbf{contrastive 1}       & 16.89 $\pm$ 0.18 & 18.34 $\pm$ 0.59 \\\midrule
\multirow{2}{*}{\textbf{\begin{tabular}[c]{@{}l@{}}Que-Es\\(updated)\end{tabular}}} & \textbf{contrastive 1} & 16.51 $\pm$ 1.12 & 14.98 $\pm$ 0.16 \\
& \textbf{contrastive 2} & 16.56 $\pm$ 0.30 & 15.66 $\pm$ 0.60
\\\bottomrule
\end{tabular}
}
\caption{Statistics (BLEU average and standard deviation) for the submitted models which have 3 runs with different seeds. The \textit{Taq-Fr} and \textit{Que-Es} BLEU scores are respectively over the IWSLT 2022 test set and the IWSLT 2023 validation set. 
}
\label{tab:results:low:stats}
\end{table}

%% file: tables/zeroshot_langid.tex
\begin{table}[]
\resizebox{\columnwidth}{!}{
\begin{tabular}{cccc|cc}\toprule
\textbf{\begin{tabular}[c]{@{}c@{}}Adapter \\ Size\end{tabular}} & \textbf{\begin{tabular}[c]{@{}c@{}}Encoder \\ Adapters\end{tabular}} & \textbf{\begin{tabular}[c]{@{}c@{}}Decoder \\ Adapters\end{tabular}} & \textbf{\begin{tabular}[c]{@{}c@{}}Taq-En\\ Lang ID\end{tabular}} & \textbf{\begin{tabular}[c]{@{}c@{}}Taq-Ko\\ Lang ID\end{tabular}} &
\textbf{\begin{tabular}[c]{@{}c@{}}Hangul\\ Percentage\end{tabular}}\\\midrule
 64 & \ding{51} & \ding{51} & 100\% & 97\% & 88\%  \\
 128 & \ding{51} & \ding{51} & 99\% & 84\% & 59\%  \\
64 & \ding{51} & \ding{55} & 100\% & 100\% & 95\% \\
\ding{55} & \ding{55} & \ding{55} & 100\% & 100\% & 96\% \\\midrule
 \ding{55} & \ding{55} & \ding{55} & 100\% & 100\% & 93\% \\
\bottomrule
\end{tabular}
}
\caption{Percentage of output sentences in the correct language according to the NLLB language ID \cite{costa2022no}. The last column shows the percentage of output characters that are in the Korean alphabet.}
\label{tab:zeroshot:langid}
\end{table}

%% file: tables/vocab_filtering.tex
\begin{table}[]
\scriptsize
\centering
\resizebox{\columnwidth}{!}{
\begin{tabular}{c|c|c|c|c|c}\toprule
\multirow{2}{*}{\textbf{Train vocab}} & \multirow{2}{*}{\textbf{Inference vocab}} & \textbf{Inference} & \textbf{Taq-Fr} & \textbf{Fr-En} & \multirow{2}{*}{\textbf{Speed}} \\
& & \textbf{params} & \textbf{BLEU} & \textbf{BLEU} & \\
\midrule
\multirow{2}{*}{\underline{Full (256k)}} & \underline{Full (256k)} & \underline{1.38B} & \underline{19.1} & \underline{\textbf{36.6}} & \underline{12.5$\times$} \\
& Filtered (35k) & 1.19B & 18.9 & 35.8 & \textbf{13.0$\times$} \\
Filtered (35k) & Filtered (35k) & 1.19B & \textbf{20.0} & 35.5 & \textbf{13.0$\times$} \\
\bottomrule 
\end{tabular}
}
\caption{Speech Translation performance on the IWSLT 2022 \textit{Taq-Fr} and mTEDx \textit{Fr-En} test sets of our contrastive \textit{Taq-Fr} submission (non-ensemble version of our primary submission) with several vocabulary filtering strategies: no filtering (first row, corresponds to our submission); inference-time filtering (second row); or training-time filtering~(third row). See Table~\ref{tab:ablation_study} for an explanation of the ``speed'' column.}
\label{tab:vocab_filtering}
\end{table}

%% file: tables/data_setup_check.tex
\begin{table*}[]
\resizebox{\textwidth}{!}{
\begin{tabular}{lccc|c|cccc|cccccccc}\toprule
                      & \multicolumn{3}{c|}{\textbf{IWSLT 2023}}                & \textbf{TED-LIUM v2} & \multicolumn{4}{c|}{\textbf{mTEDx ASR}}                                    & \multicolumn{8}{c}{\textbf{mTEDx ST}}                                                                                                                 \\
\textbf{Submission}   & \textbf{Taq-Fr} & \textbf{Que-Es} & \textbf{Que-Que} & \textbf{En-En}     & \textbf{Fr-Fr} & \textbf{Es-Es} & \textbf{It-It} & \textbf{Pt-Pt} & \textbf{Fr-En} & \textbf{Fr-Es} & \textbf{Es-Fr} & \textbf{Es-En} & \textbf{Fr-Pt} & \textbf{Pt-En} & \textbf{Es-It} & \textbf{Es-Pt} \\\midrule
Taq-Fr primary       & $$ \ding{51} $$                &     $$ \ding{55} $$   &   $$ \ding{55} $$      &  $$ \ding{51} $$        &  $$ \ding{51} $$    &  $$ \ding{51} $$    & $$ \ding{55} $$ & $$ \ding{55} $$ &  $$ \ding{51} $$    &  $$ \ding{51} $$    &  $$ \ding{51} $$    &  $$ \ding{51} $$    & $$ \ding{55} $$ & $$ \ding{55} $$ & $$ \ding{55} $$& $$ \ding{55} $$     \\
Taq-Fr contrastive 1 &  $$ \ding{51} $$    & $$ \ding{55} $$ & $$ \ding{55} $$ &  $$ \ding{51} $$        &  $$ \ding{51} $$    &  $$ \ding{51} $$    & $$ \ding{55} $$ & $$ \ding{55} $$ &  $$ \ding{51} $$    &  $$ \ding{51} $$    &  $$ \ding{51} $$    &  $$ \ding{51} $$    & $$ \ding{55} $$ & $$ \ding{55} $$ & $$ \ding{55} $$& $$ \ding{55} $$     \\
Taq-Fr contrastive 2 &  $$ \ding{51} $$    &  $$ \ding{51} $$    &  $$ \ding{51} $$    &  $$ \ding{51} $$        &  $$ \ding{51} $$    &  $$ \ding{51} $$    & $$ \ding{55} $$ & $$ \ding{55} $$ &  $$ \ding{51} $$    &  $$ \ding{51} $$    &  $$ \ding{51} $$    &  $$ \ding{51} $$    & $$ \ding{55} $$ & $$ \ding{55} $$ & $$ \ding{55} $$& $$ \ding{55} $$     \\\midrule
Que-Es primary       &  $$ \ding{51} $$    &  $$ \ding{51} $$    &  $$ \ding{51} $$    &  $$ \ding{51} $$        &  $$ \ding{51} $$    &  $$ \ding{51} $$    & $$ \ding{55} $$ & $$ \ding{55} $$ &  $$ \ding{51} $$    &  $$ \ding{51} $$    &  $$ \ding{51} $$    &  $$ \ding{51} $$    & $$ \ding{55} $$ & $$ \ding{55} $$ & $$ \ding{55} $$& $$ \ding{55} $$     \\
Que-Es contrastive 1  &  $$ \ding{51} $$    &  $$ \ding{51} $$    &  $$ \ding{51} $$    &  $$ \ding{51} $$        &  $$ \ding{51} $$    &  $$ \ding{51} $$    & $$ \ding{55} $$ & $$ \ding{55} $$ &  $$ \ding{51} $$    &  $$ \ding{51} $$    &  $$ \ding{51} $$    &  $$ \ding{51} $$    & $$ \ding{55} $$ & $$ \ding{55} $$ & $$ \ding{55} $$& $$ \ding{55} $$     \\
Que-Es contrastive 2 &  $$ \ding{51} $$    &  $$ \ding{51} $$    &  $$ \ding{51} $$    &  $$ \ding{51} $$        &  $$ \ding{51} $$    &  $$ \ding{51} $$    & $$ \ding{55} $$ & $$ \ding{55} $$ &  $$ \ding{51} $$    &  $$ \ding{51} $$    &  $$ \ding{51} $$    &  $$ \ding{51} $$    & $$ \ding{55} $$ & $$ \ding{55} $$ & $$ \ding{55} $$& $$ \ding{55} $$     \\\midrule
IWSLT 2021 setup      &  $$ \ding{51} $$    &  $$ \ding{51} $$    &  $$ \ding{51} $$    &  $$ \ding{51} $$        &  $$ \ding{51} $$    &  $$ \ding{51} $$    &  $$ \ding{51} $$    &  $$ \ding{51} $$    &  $$ \ding{51} $$    &  $$ \ding{51} $$    &  $$ \ding{51} $$    &  $$ \ding{51} $$    &  $$ \ding{51} $$    &  $$ \ding{51} $$    &  $$ \ding{51} $$    & $$ \ding{51} $$   \\\bottomrule         
\end{tabular}}
\caption{Extensive list of datasets used for training~(\ding{51}) each system presented in this paper.}
\label{tab:data:check}
\end{table*}

%% file: tables/models_speech2.tex
\begin{table*}[h]
\resizebox{\textwidth}{!}{
\begin{tabular}{ll}
\toprule
\textbf{Model}                      & \textbf{URL}                                                                \\\midrule
\textbf{mHuBERT-Tamasheq}           & Unavailable                                                                 \\
\textbf{Tamasheq}                 & \url{https://huggingface.co/LIA-AvignonUniversity/IWSLT2022-tamasheq-only}           \\
\textbf{Niger-Mali}                 & \url{https://huggingface.co/LIA-AvignonUniversity/IWSLT2022-Niger-Mali}           \\
\textbf{XLSR-53}                    & \url{https://github.com/facebookresearch/fairseq/tree/main/examples/wav2vec}      \\
\textbf{XLS-R large and xlarge}     & \url{https://github.com/facebookresearch/fairseq/tree/main/examples/wav2vec/xlsr} \\
\bottomrule
\end{tabular}}
\caption{Downloading sources for the speech representation models checkpoints used in our experiments. 
} 
\label{tab:models:speechurl}
\end{table*}

%% file: tables/stacked_layers.tex
\begin{table*}[]
\centering
\begin{tabular}{c|c|c|c|c|c|c|c}\toprule
\textbf{Stacked} & \textbf{FT} & \multirow{2}{*}{\textbf{Adapters}} & \textbf{Total} & \textbf{Trained} & \textbf{Taq-Fr} & \textbf{Fr-En} & \multirow{2}{*}{\textbf{Speed}} \\
\textbf{layers} & \textbf{layers} & & \textbf{params} & \textbf{params} & \textbf{BLEU} & \textbf{BLEU} & \\
\midrule
1 & 0 & enc+dec (64) & 1.40B & 28M & \textbf{19.2} & 35.0 & 12.0$\times$ \\ 
1 & 0 & none & 1.39B & 22M & 17.9 & 33.8 & 12.2$\times$ \\ 
0 & 1 & enc+dec (64) & 1.38B & 28M & 18.2 & 35.1 & 12.0$\times$ \\ 
0 & 1 & none & 1.37B & 22M & 17.5 & 33.3 & 12.6$\times$ \\ 
\hline
2 & 0 & enc+dec (64) & 1.42B & 49M & \textbf{19.2} & 35.1 & 11.9$\times$ \\ 
2 & 0 & none & 1.41B & 43M & 18.4 & 35.0 & 12.5$\times$ \\ 
0 & 2 & enc+dec (64) & 1.38B & 49M & 19.0 & 36.2 & 12.0$\times$ \\ 
\hline
0 & 3 & enc+dec (64) & \underline{1.38B} & \underline{70M} & \underline{19.1} & \underline{\textbf{36.6}} & \underline{12.5$\times$} \\\bottomrule 
\end{tabular}
\caption{Training stacked layers~(i.e. adding and training new bottom encoder layers) versus fine-tuning the existing bottom layers; with or without adapters. The other hyper-parameters are identical to our constrastive submission~(underlined scores).}
\label{tab:stacked_layers}
\end{table*}

%% file: tables/ablation_study.tex
\begin{table*}[]
\scriptsize
\resizebox{\textwidth}{!}{
\begin{tabular}{c|c|c|c|c|c|c|c|c|c}\toprule
\multirow{2}{*}{\textbf{Speech features}} & \multirow{2}{*}{\textbf{MT model}} & \textbf{Conv.} & \textbf{FT} & \multirow{2}{*}{\textbf{Adapters}} & \textbf{Total} & \textbf{Trained} & \textbf{Taq-Fr} & \textbf{Fr-En} & \multirow{2}{*}{\textbf{Speed}} \\
& & \textbf{layers} & \textbf{layers} & & \textbf{params} & \textbf{params} & \textbf{BLEU} & \textbf{BLEU} & \\
\midrule
Tamasheq (layer 11) & \multirow{9}{*}{NLLB 1.3B} & \multirow{9}{*}{1} & \multirow{9}{*}{3} & \multirow{9}{*}{enc+dec (64)} & 1.38B & 70M & 16.8 & 32.5 & 11.6$\times$ \\ 
Tamasheq (layer 8) & & & & & 1.38B & 70M & 19.3 & 31.6 & 12.0$\times$ \\ 
mHuBERT-Taq (layer 11) & & & & & 1.38B & 70M & 16.4 & 37.1 & 12.1$\times$ \\ 
mHuBERT-Taq (layer 8) & & & & & 1.38B & 70M & 16.2 & 36.7 & 12.1$\times$ \\ 
Niger-Mali (layer 11) & & & & & 1.38B & 70M & 16.6 & 34.6 & 11.8$\times$ \\ 
Niger-Mali (layer 8) & & & & & \underline{1.38B} & \underline{70M} & \underline{19.1} & \underline{36.6} & \underline{12.5$\times$} \\ 
XLSR-53 (layer 18) & & & & & 1.38B & 70M & 15.9 & 38.0 & 12.4$\times$ \\ 
XLS-R L (layer 18) & & & & & 1.38B & 70M & 16.8 & \textbf{39.4} & 12.7$\times$ \\ 
XLS-R XL (layer 46) & & & & & 1.38B & 70M &  15.4 & 37.4 & 11.7$\times$  \\ 
\hline
\multirow{4}{*}{Niger-Mali (layer 8)} & mBART (600M) & \multirow{4}{*}{1} & \multirow{4}{*}{3} & \multirow{4}{*}{enc+dec (64)} & 0.61B & 41M & 16.3 & 28.9 & 22.9$\times$ \\ 
& NLLB (600M) & & & & 0.62B & 41M & 18.0 & 32.5 & 24.2$\times$ \\ 
& NLLB (1.3B) & & & & \underline{1.38B} & \underline{70M} & \underline{19.1} & \underline{36.6} & \underline{12.5$\times$} \\ 
& NLLB (3.3B) & & & & 3.36B & 165M & 19.3 & 37.3 & 4.5$\times$ \\ 
\hline
\multirow{4}{*}{Niger-Mali (layer 8)} & \multirow{4}{*}{NLLB 1.3B} & 3 & \multirow{4}{*}{3} & \multirow{4}{*}{enc+dec (64)} & 1.38B & 70M & 18.5 & 33.4 & \textbf{25.5$\times$} \\ 
& & 2 & & & 1.38B & 70M & 19.4 & 35.4 & 19.5$\times$ \\ 
& & 1 & & & \underline{1.38B} & \underline{70M} & \underline{19.1} & \underline{36.6} & \underline{12.5$\times$} \\ 
& & 0 & & & 1.38B & 70M & \textbf{19.6} & 34.4 & 7.1$\times$ \\ 
\hline
\multirow{5}{*}{Niger-Mali (layer 8)} & \multirow{5}{*}{NLLB 1.3B} & \multirow{5}{*}{1} & 24 & \multirow{5}{*}{enc+dec (64)} & 1.37B & 508M & 16.7 & 30.7 & 11.9$\times$ \\ 
& & & 4 & & 1.38B & 91M & 19.6 & 36.8 & 12.3$\times$ \\ 
& & & 3 & & \underline{1.38B} & \underline{70M} & \underline{19.1} & \underline{36.6} & \underline{12.5$\times$} \\ 
& & & 2 & & 1.38B & 49M & 19.0 & 36.2 & 12.0$\times$ \\ 
& & & 1 & & 1.38B & 28M & 18.2 & 35.1 & 12.0$\times$ \\ 
\hline
\multirow{7}{*}{Niger-Mali (layer 8)} & \multirow{7}{*}{NLLB 1.3B} & \multirow{7}{*}{1} & \multirow{2}{*}{1} & enc (64) & 1.37B & 25M & 19.1 & 34.2 & 12.4$\times$ \\ 
& & & & none & 1.37B & \textbf{22M} & 17.5 & 33.3 & 12.6$\times$ \\ 
\cline{4-5}
& & & \multirow{5}{*}{3} & enc+dec (256) & 1.40B & 88M & 18.8 & 35.8 & 12.2$\times$ \\ 
& & & & enc+dec (128) & 1.38B & 76M & 19.2 & 36.3 & 12.1$\times$ \\ 
& & & & enc+dec (64) & \underline{1.38B} & \underline{70M} & \underline{19.1} & \underline{36.6} & \underline{12.5$\times$} \\ 
& & & & enc (64) & 1.37B & 67M & 19.3 & 35.7 & 12.7$\times$ \\ 
& & & & none & 1.37B & 64M & 18.3 & 35.6 & 13.1$\times$ \\ 
\bottomrule
\end{tabular}
}
\caption{Ablation study on \textit{Taq-Fr} ST, with various speech feature extractors, pre-trained MT models used for initialization, and trained parameters. The total parameter counts do not include the parameters of the speech feature extractors. The BLEU scores reported are on the IWSLT 2022 \textit{Taq-Fr} and mTEDx \textit{Fr-En} test sets. The speed metric is relative to real time~(i.e., seconds in the test set divided by seconds spent decoding) and does not include feature extraction time. It is obtained by decoding the \textit{Taq-Fr} test set on a single T4 with a batch size of 10 utterances (averaged over 3 decoding runs). The underlined numbers all correspond to the same model, which is our first contrastive submission to the task (the non-ensemble version of our primary submission).
All of these models are trained with the same data~(see Table~\ref{tab:data:check}) 
and early stopping is done based on \textit{Taq-Fr} valid BLEU scores.
The numbers inside parentheses in the \textit{Adapters} column correspond to the bottleneck dimension of the trained adapter modules. Adapters are not added in the encoder layers that are being fine-tuned. These models took between 15 and 47\,h each to train on 4 V100 GPUs, with an average training time of 26\,h.}
\label{tab:ablation_study}
\end{table*}

\begin{figure}
    \centering
    \includegraphics[width=0.5\textwidth]{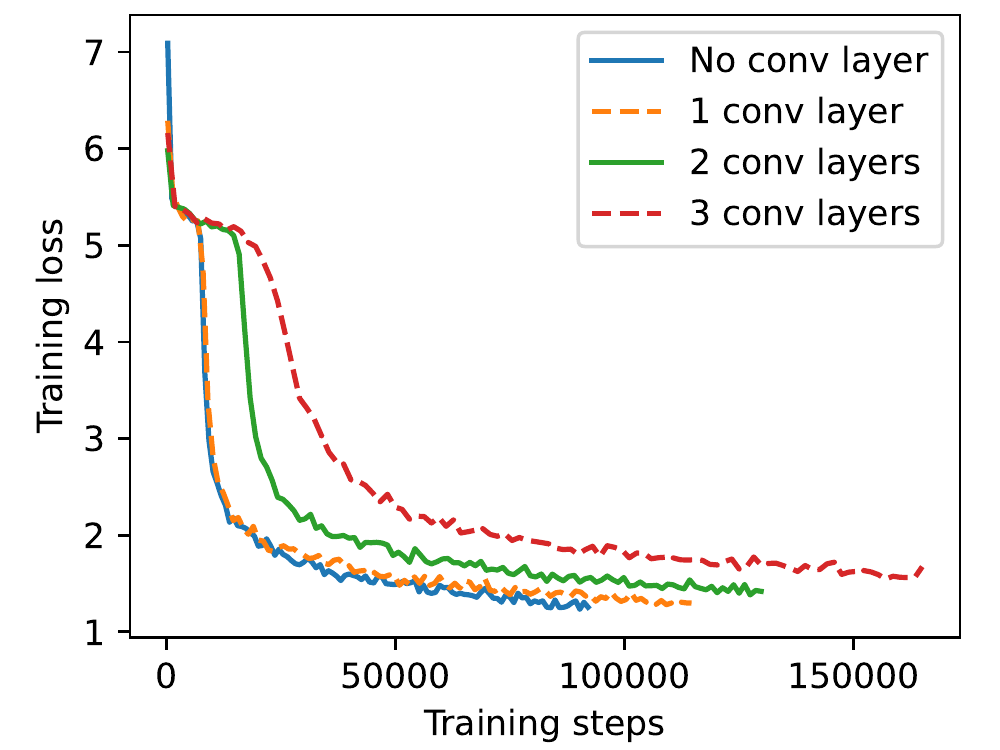}
    \includegraphics[width=0.5\textwidth]{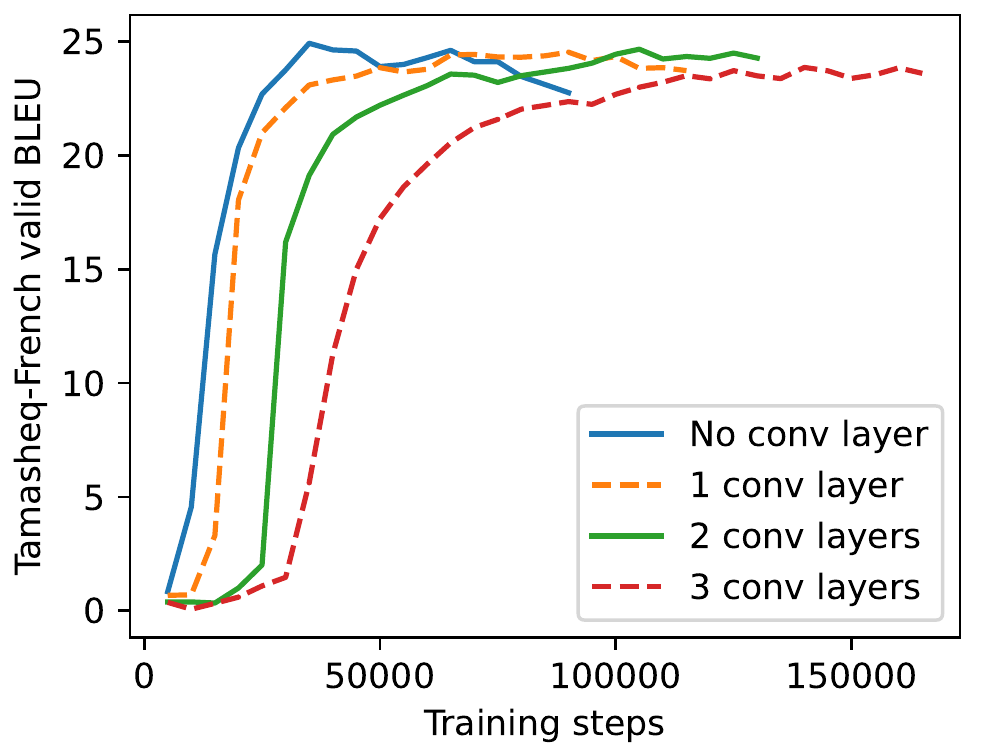}
    \caption{Training loss and \textit{Taq-Fr} validation BLEU of variants of our \emph{contrastive 1} model, that have 0 to 3 convolutional layers (1 by default).}
    \label{fig:bleu_by_conv}
\end{figure}

\begin{figure}
    \centering
    \includegraphics[width=0.5\textwidth]{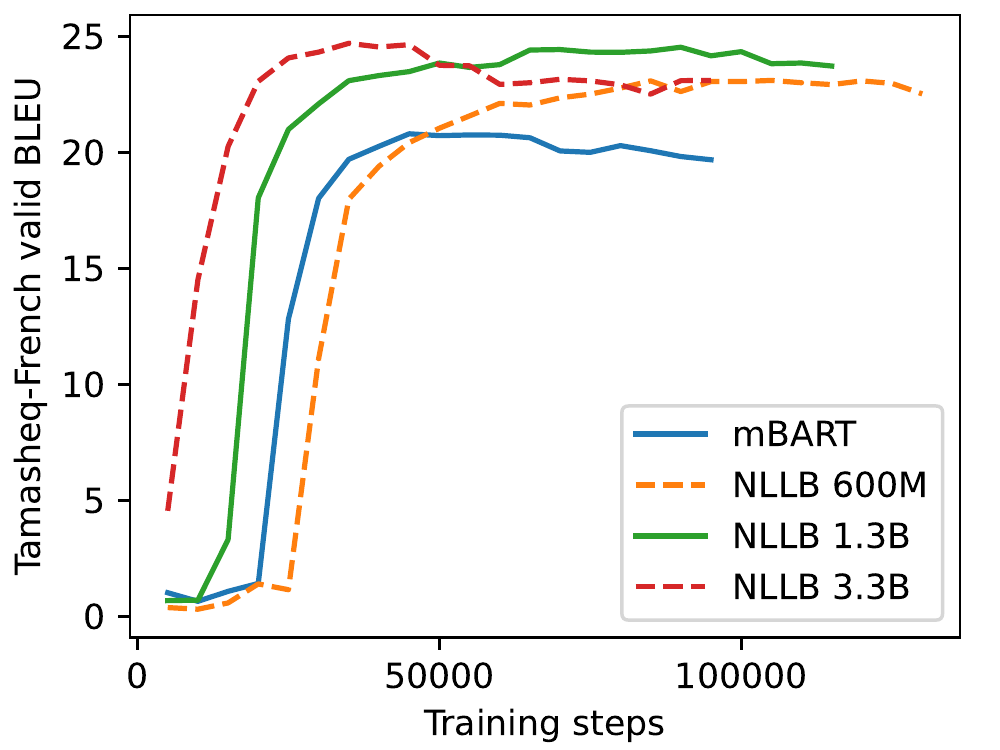}
    \caption{\textit{Taq-Fr} validation BLEU of variants of our \emph{contrastive 1} model that are initialized with various MT models (NLLB 1.3B by default).}
    \label{fig:bleu_by_mt_model}
\end{figure}

\begin{figure}
    \centering
    \includegraphics[width=0.5\textwidth]{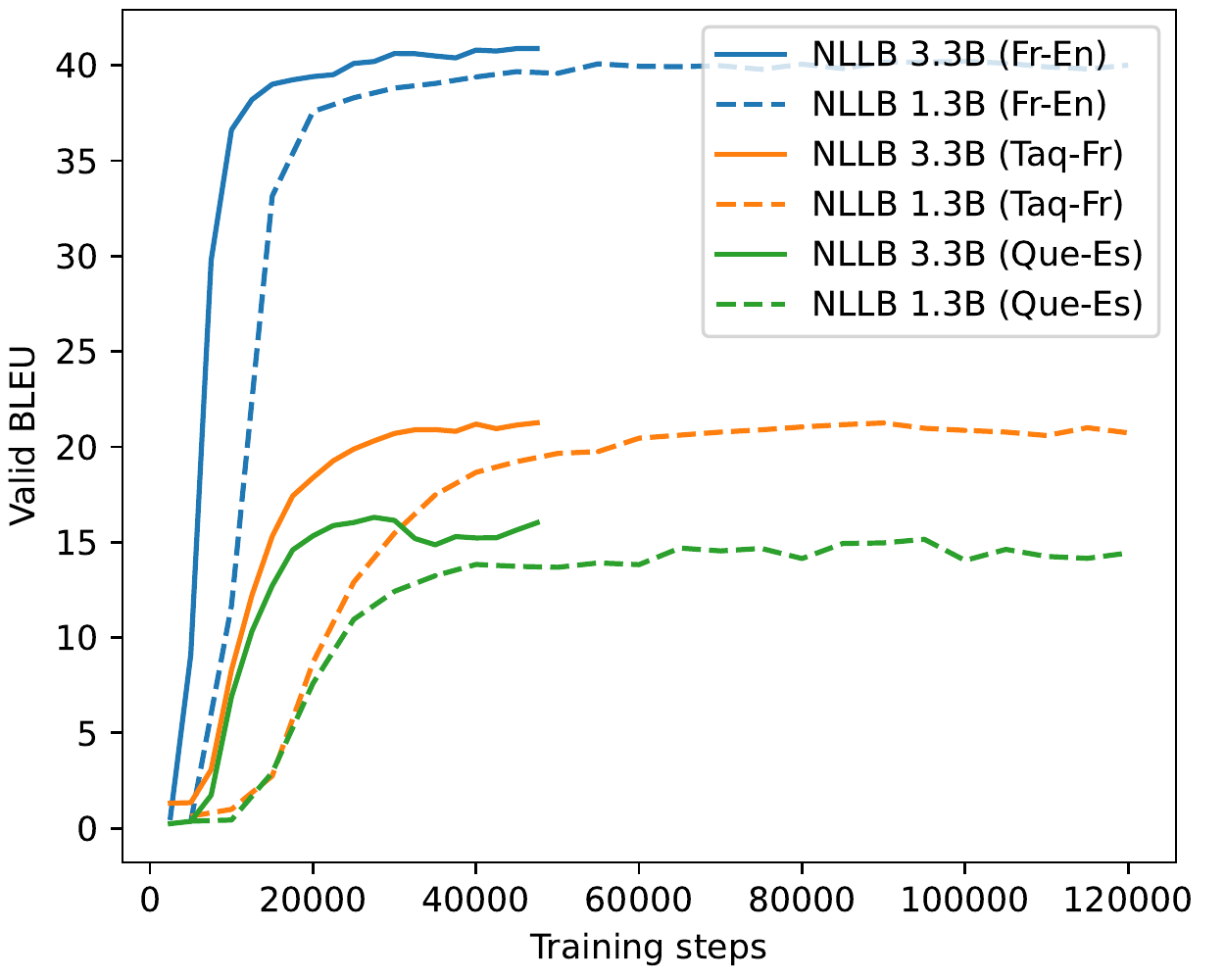}
    \caption{Validation BLEU by language direction (\textit{Fr-En}, \textit{Taq-Fr} and \textit{Que-Es}) of a multilingual model (XLS-R + NLLB 1.3B) which includes both Tamasheq and Quechua (our \emph{updated constrastive~1} submission).}
    \label{fig:bleu_by_lang}
\end{figure}


\begin{table*}[]
\resizebox{\textwidth}{!}{
\begin{tabular}{c|c|c|cccc|ccc}\toprule
\multirow{2}{*}{\textbf{Task}} & \multirow{2}{*}{\textbf{Model}} & \multirow{2}{*}{\textbf{Adapters}} & \multicolumn{4}{c|}{\textbf{Training directions}} & \multicolumn{3}{c}{\textbf{Zero-shot directions}} \\
& & & Es-En & Fr-En & Fr-Es & Pt-En & Pt-Es & It-En & It-Es \\
\hline
ST & NLLB 3.3B & enc+dec & 44.0 & \textbf{39.9} & 38.3 & \textbf{33.1} & 38.1 & 29.3 & 36.9 \\
\hline
\multirow{4}{*}{ST} & \multirow{4}{*}{NLLB 1.3B} & enc+dec & 43.7 & 39.4 & 38.0 & 31.5 & 35.9 & 28.9 & 35.0 \\ 
& & none & 36.7 & 35.0 & 31.7 & 23.8 & 30.5 & 25.2 & 31.3 \\
& & enc & 41.4 & 38.3 & 36.0 & 30.8 & 36.2 & 26.2 & 35.1 \\
& & dec & 39.1 & 38.2 & 33.1 & 26.9 & 31.9 & 27.9 & 32.9 \\


\bottomrule
\toprule
MT & NLLB 3.3B & none & 47.4 & 39.5 & 39.2 & 39.8 & 48.6 & 34.0 & 42.4 \\ 
\hline
\multirow{4}{*}{MT} & \multirow{4}{*}{NLLB 1.3B} & none & 47.9 & 38.9 & 39.6 & 39.8 & 48.5 & 33.8 & 41.9 \\ 
& & enc+dec & 50.2 & 40.7 & 42.2 & 42.1 & 51.0 & 37.6 & 45.2 \\ 
& & enc & 49.9 & 41.3 & 42.6 & 41.9 & 50.6 & 36.5 & 44.9 \\ 
& & dec & 48.8 & 39.2 & 41.0 & 41.1 & 49.7 & 35.6 & 43.9 \\ 
\hline
MT & NLLB 1.3B (DA) & enc+dec & \textbf{51.3} & \textbf{43.2} & \textbf{45.2} & \textbf{44.7} & \textbf{53.2} & \textbf{37.8} & \textbf{47.1} \\ 
\bottomrule
\end{tabular}
}
\caption{\textbf{Top half:} Speech translation BLEU scores on the IWSLT 2021 test sets, when deactivating encoder adapters, decoder adapters, or both in an ST model at inference time. The ST model is the same one as in Table~\ref{tab:iwslt_2021_setting}, trained with encoder and decoder adapters. \textbf{Bottom half:} Text-to-text MT BLEU scores when using the ST adapters in the initial model and disabling the ST bottom layers and convolutions.}
\label{tab:domain_transfer}
\end{table*}